\documentclass[a4paper, 10pt, conference]{ieeeconf}      

\IEEEoverridecommandlockouts                              
\usepackage{amsmath} 
\usepackage{amssymb}  
\usepackage{cite}
\usepackage{threeparttable}
\usepackage{stfloats}
\usepackage{comment}
\usepackage{gensymb}

\title{\LARGE \bf
Guided Policy Search Model-based Reinforcement Learning\\
for Urban Autonomous Driving
}

\author{Zhuo Xu, Jianyu Chen, and Masayoshi Tomizuka
\thanks{{$^\dagger$}Z. Xu, J. Chen and M. Tomizuka are with the Department of Mechanical
Engineering, University of California, Berkeley, CA 94720 USA (e-mail:
zhuoxu@berkeley.edu, jianyuchen@berkeley.edu, tomizuka@berkeley.edu).}}

\usepackage{algorithm,algpseudocode}
\usepackage[algo2e]{algorithm2e} 
\usepackage{graphicx}\usepackage{multirow}
\usepackage{url}
\begin{document}

\maketitle
\thispagestyle{empty}
\pagestyle{empty}

\begin{abstract}
In this paper, we continue our prior work on using imitation learning (IL) and model free reinforcement learning (RL) to learn driving policies for autonomous driving in urban scenarios, by introducing a model based RL method to drive the autonomous vehicle in the Carla urban driving simulator. Although IL and model free RL methods have been proved to be capable of solving lots of challenging tasks, including playing video games, robots, and, in our prior work, urban driving, the low sample efficiency of such methods greatly limits their applications on actual autonomous driving. In this work, we developed a model based RL algorithm of guided policy search (GPS) for urban driving tasks. The algorithm iteratively learns a parameterized dynamic model to approximate the complex and interactive driving task, and optimizes the driving policy under the nonlinear approximate dynamic model. As a model based RL approach, when applied in urban autonomous driving, the GPS has the advantages of higher sample efficiency, better interpretability, and greater stability. We provide extensive experiments validating the effectiveness of the proposed method to learn robust driving policy for urban driving in Carla. We also compare the proposed method with other policy search and model free RL baselines, showing 100x better sample efficiency of the GPS based RL method, and also that the GPS based method can learn policies for harder tasks that the baseline methods can hardly learn.
\end{abstract}

\section{Introduction}

Autonomous driving in urban scenario is an extremely complex task requiring extensive decision making within dense surrounding environment. The currently popular non-learning rule-based or model-based methods often fail to accomplish satisfactory driving performance in such high complex driving tasks due to its low accuracy introduced by the biased human-designed heuristics and lack of generality. Therefore, in addition to the low dimensional control methods, higher dimensional learning based approaches are favorable to provide higher level decision making and motion planning guidance.

The possibility of using the high dimensional learning based methods to drive the autonomous cars is enabled by the recent advances of the machine learning technology. Massive data and computation power have advanced the performance of the learning methods in various fields of computer vision, data science, robotics and autonomous driving. In the field of autonomous diving, the most popular and studied learning based methods are imitation learning (IL) based method and model free reinforcement learning (RL) methods.

IL learning methods involve first collecting a great amount of data of the autonomous vehicles' on board observation and some expert driving control commands. The major part of the IL method is essentially a supervised learning to generate a driving policy to map the observation input to the control output, cloning the expert's driving behavior. However, the IL methods suffer from a number of shortcomings: 1) it requires collecting a large amount of driving data, which is costly and time consuming. 2) the training dataset is biased compared to real world driving, since the expert drivers generally do not provide data for dangerous situations. 3) IL essentially clones the human driver behavior, and cannot exceed the human performance.

Compared with the IL methods, RL methods are more suitable for training autonomous driving policies in that it can have the simulated agent to explore the driving scenarios without human guidance. Therefore, the policies trained are able to learn cases that are not in the IL training set, and can potentially exceed performance of the human drivers. Within the class of RL methods, the more intuitive and studied methods are model free RL methods. In our prior work, we have applied a class of baseline model free RL methods to learn driving policies in Carla.

However, model free RL methods suffer from low sample efficiency and lack of interpretability, which limits its application in real world driving. Since RL policies are learned in simulators, which generally cannot model the real world driving scenarios 100\%, the major concern of the deployment of the RL policies in the real world is the transfer learning to overcome the model error between the simulation and the real world, or the so called reality gap. Model free RL methods, which directly optimize the driving policy in an end to end manner, requires a huge amount of driving episodes for the finetuning of the driving policy, which is time and computation consuming. Moreover, the on board deployment requires massive finetuning as well, and this can be extremely dangerous. What makes it worse is the low interpretable level of the end to end policy, and that the RL agent randomly explores the environment without any guidance. These can lead to computation waste, and can limit the policy to local optimum, resulting in failing to learn optimal policies in constrained tasks (for example, driving tasks with obstacles).

In this paper, we propose to model based RL methods for urban autonomous driving. Model based RL lies in the intersection of model based planning control and RL, which iteratively learns the model of the complex environment, and optimizes the control policy base on the understanding of the environment dynamics. Model based RL methods have been proved successful in dealing with real world robotic manipulation and aerial vehicle navigation, but the applications using model based RL in autonomous driving has lag behind, given most basic model based RL method lack the capacity of solving complicated driving problems, while more powerful algorithms developed in recent years have hardly been applied to autonomous driving. We modify the guided policy search model based RL approach to learn the autonomous driving policy, and implement and compare other policy search and model free RL baselines. Final results show that our model based method can learn a robust driving policy in the urban scenario, have much better sample efficiency compared to the model free methods, and is capable to learn to drive in more challenging tasks that the prior methods can hardly solve.

The rest of this paper is organized as follows. The related work is summarized in Section II. In Section III, the problem setup and our modified guided policy search model based RL algorithm designed especially for the autonomous driving task is described. Then in Section IV, the setting of the Carla urban autonomous driving simulator, together with the other model based RL method and baseline model free RL algorithms, are illustrated. In Section V, simulation results are presented and evaluated to show the effectiveness and advantages of the proposed method, followed by the conclusion in Section VI.

\section{Related Work}
Most literature on using deep learning to solve autonomous driving problems is based on IL, which essentially use neural network to do behavior cloning of the input-output combinations of the human expert drivers. One of the earliest attempts are the ALVINN \cite{pomerleau1989alvinn} and \cite{muller2006off}, and more recent works include \cite{nvidia2016end}, \cite{xu2016end}. Waymo has also applied IL to learn a urban driving policy from huge human driving data \cite{bansal2018chauffeurnet}. The progress over the years is mainly on using larger dataset and more sophisticated neural network structures. However, since the policies are still trained in a supervised manner, they are inevitably limited by the dataset. Our previous work sought to obtain a more stable policy by first learning a future trajectory and then have a low level controller to track the trajectory \cite{chen2019deep}.

Model based RL has been demonstrated to be capable of solving many challenging tasks including in control \cite{lillicrap2015continuous,mnih2016asynchronous}. Various deep RL methods including deep Q-learning and actor-critic algorithms have been applied to train lane-keeping policies with low-dimensional feature vector or image pixels as inputs \cite{sallab2016end}. Another deep Q learning application is \cite{wolf2017learning}, which learns the steer command of an autonomous vehicle in simulation with discrete action space. Wayve's researchers used more sophisticated network and representation learning to use RL to drive an autonomous vehicle in urban scenarios \cite{kendall2019learning}. Our previous work on model free RL provided a benchmark comparison of different model free RL algorithms in the Carla simulator \cite{chen2019model}.

However, as is pointed out, the model free reinforcement learning suffers from great sample inefficiency\cite{xu2017cascade, xu2019toward}. Also, since the neural network policies are trained in an end to end manner, it lacks of interpretability and can be very hard to transfer the real world vehicles \cite{xu2018zero, tang2019disturbance}. The model based or policy search based RL are more sample efficient and have been proved capable of solving a number of complicated manipulation tasks \cite{levine2016end} and aerial vehicle \cite{zhang2016learning} in both simulator and real world. In addition to the guided policy search method \cite{levine2013guided}, we also test and compare it to one of the most popular policy search method of cross entropy method \cite{de2005tutorial} to validate the effectiveness of the proposed method in urban autonomous driving.

\section{Methodology}
\subsection{Problem setup}
We model the autonomous driving task using a Markov Decision Process (MDP), whose state is defined using $s_t$, the agent interacts with the environment by taking actions $a_t$, derived using its control policy $\pi_\theta (a_t | s_t)$, and the control policy is parameterized by $\theta$. In autonomous driving, the MDP state can include the lane tracking status, speed of the autonomous vehicle, and the obstacle vehicles. The MDP action is the control command for the autonomous vehicle, corresponding to throttle, brake, and steering angle. Accepting the action of the agent, the environment would evolve in time according to the system dynamics $p(s_{t+1} | s_t, a_t)$. The system dynamics can be very complicated in general for the urban driving tasks given the dense traffic interaction. Furthermore, the obstacles movements are usually not controllable by the control commands of the autonomous vehicle.

With the system governed by the dynamics function and the control policy, the trajectory can be described as a distribution of 
\begin{align}
\pi_\theta(\tau) = p(x_0) \prod_{t=0}^{T-1} \pi_\theta(a_t|s_t) p(s_{t+1}|s_t, a_t)
\end{align}
where $\tau=\{s_0, a_0, s_1, a_1, ..., s_T, a_T\}$. The goal of the autonomous driving task is described by a cost function, $l(s_t, a_t)$, which can be penalties for the lateral and reference speed deviation of the autonomous vehicle, or for the collisions with the obstacle. The overall objective would then be to minimize the expectation $E_{\pi_\theta(\tau)}[\Sigma l(s_t, a_t)]$, which we will abbreviate as $E_{\pi_\theta} [l(\tau))]$. It is noted that the problem formulation for RL is the same as that of the optimal control, only the system dynamics $p(s_{t+1}|s_t, a_t)$ is not known, and often very complex, such that one can not precisely model using mathematical equations, but can only use parametrized models to approximate it. This brings challenges to the learning of a driving policy, but also the setting can enable the RL policy to control the vehicle under complicated vehicle dynamics, and dense traffic that is hard to model.

\subsection{Approach Summary}

\begin{figure*}[t]
\centering
\includegraphics[scale=0.36]{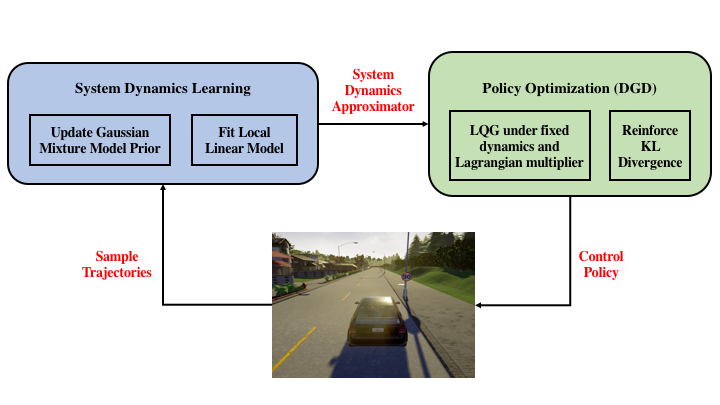}
\caption{The framework overflow of our proposed guided policy search (GPS) model based RL method. For each iteration, we first collect trajectories in the simulator using the current policy, then the trajectory samples are used to update the system dynamics approximator, which is further used as the model within the model based policy optimization.}
\label{fig:structure}
\end{figure*}

The model based RL method, as described in the Introduction Section, is a RL method that combines the strengths of both RL and optimal control. The main idea, as shown in Fig. 1, is to iteratively updates the belief of the system dynamics model and to optimize the control policy under the current dynamics approximation function. Concretely, the RL agent first explores the environment, collect trajectory samples of $\tau=\{s_0, a_0, s_1, a_1, ..., s_T, a_T\}$, and use the collected data tuples $(s_t, a_t, s_{t+1})$ to learn a dynamics function $p(s_{t+1}|s_t, a_t)$. That is a supervised learning regression process, optimizing:

\begin{align}
\min_{p(s_{t+1}|s_t, a_t)} p(\tau)
\end{align}

Then we apply trajectory optimization methods to learn the optimal policy $\pi_\theta(a_t|s_t)$, which can control the agent to obtain optimal reward $E_{\pi_\theta} [l(\tau))]$ under the learned dynamics $p(s_{t+1}|s_t, a_t)$, and then iterate. One can also use IL-alike method to apply supervised learning to learn a neural network policy that clone the behavior of $\pi_\theta(a_t|s_t)$ For our autonomous driving problem, we made the modification by focusing on the update of the system dynamics model and the control policy.

Since the system dynamics and the control policies can be very complicated, they can only be approximated using high dimensional parameterized models, such as Gaussian mixture models or neural networks. However, it can take a number of episodes for the training of such parameterized models. In order to get high sample efficiency, we adopt the idea of local models, and apply the time-varying linear-Gaussian models to approximate the local behavior of the system dynamics and the control policy, as:

\begin{align}
p(s_{t+1}|s_t, a_t) = \mathcal{N} (A_t s_t+B_t a_t +f_t, F_t)
\end{align}
\begin{align}
\pi_\theta(a_t|s_t) =  \mathcal{N} (K_t s_t+k_t, C_t)
\end{align}

The local linear models have the benefit of being able to be learned very efficiently with a small number of samples, but they can only describe the properties of the nonlinear functions, and thus for each policy optimization iteration, the new trajectory produced by the new policy, denoted $\tau$, shall not be similar to the old trajectory samples we used to learn dynamics $p(s_{t+1}|s_t, a_t)$. We denote the old trajectories using $\hat{\tau}$. We adopt the KL-divergence to describe the change of the trajectory distributions, then the policy optimization process can be modeled using the optimization problem below.

\begin{align}
\min_\theta E_{\pi_\theta} [l(\tau))]
\end{align}
\begin{align}
s.t. D_{KL} (p(\tau)\|p(\hat{\tau}))<\epsilon
\end{align}

In the two following subsections, we illustrate how to efficiently do system dynamics learning (2) and policy optimization (5-6).

\subsection{System Dynamics Learning}
The goal of the system dynamics learning is not only to learn a precise local linear function (3), but also to learn it highly sample efficiently. Therefore, we adopt a global model as the prior, which evolves throughout the whole model based RL lifetime, and fit the local linear dynamics to it at each iteration. The global prior itself does not itself need to be a good approximation, it only needs to capture the major property of the system dynamics, such as to increase the regression sample efficiency.

In the case of autonomous driving, since there are a series of different driving patterns and within each driving pattern, the dynamics model follows similar property. Therefore, we adopt the Gaussian mixture model (GMM) to serve as the nonlinear prior model, with each mixture element serving as prior for one driving pattern. Under this setting, each tuple sample $(s_t, a_t, s_{t+1})$ is first assigned to a pattern, and then used to update the mixture element. This process is a typical Expectation Maximization (EM) process used to train a GMM.

Finally, at each iteration, we fit the current episode of data $(s_t, a_t, s_{t+1})'s$ to the GMM, incorporating a normal-inverse-Wishart prior. The local lineal dynamics $p(s_{t+1}|s_t, a_t)$is derived by conditioning the Gaussian on $(s_t, a_t)$.

\subsection{Policy Optimization}

In order to solve for the policy optimization (5-6), we incorporate a popular gradient based method for constrained optimization, the dual gradient descent (DGD), which is summarized in algorithms 1. The main idea of the DGD is to first minimize the Lagrangian function under fixed Lagrangian multiplier $\lambda$, and then increase the $\lambda$ penalty if the constrained is violated, so that more emphasis is placed on the constraint term in the Lagrangian function in the next iteration. We first write out the Lagrangian

\begin{align}
L(\theta, \lambda) = \sum_{t=0}^{T-1} E_{\pi_\theta} [l(s_t, a_t) + \lambda (D_{KL}(p(\tau)\|p(\hat{\tau}))-\epsilon)]
\end{align}

Consider
\begin{align}
D_{KL}(p(\tau)\|p(\hat{\tau})) = E_{\pi_\theta} [log(p(\tau))-log(p(\hat{\tau}))]
\end{align}

We can reformulate the minimization of the Lagrangian function to be some trajectory optimization problem with regard to some augmented cost function $c(s_t, a_t) = l(s_t, a_t)/\lambda - log(p_{\hat{\tau}}(a_t, s_t))$. That is

\begin{align}
\min_{\pi_\theta} \sum_{t=0}^{T-1} c(s_t, a_t)
\end{align}

Since we can directly compute the cost function $c(s_t, a_t)$ and its derivatives, we can then solve the trajectory optimization problem using LQG. The LQG is omitted for the sake of space, and the readers are referred to [] for the detailed solution. After the Lagrangian is optimized under a fixed $\lambda$, in the second step of DGD, $\lambda$ is updated using the function below with step size $\alpha$, and the DGD loop is closed.

\begin{align}
\lambda \leftarrow \lambda+\alpha(D_{KL}(p(\tau)\|p(\hat{\tau}))-\epsilon))
\end{align}

\begin{algorithm}[H]
\SetAlgoLined
Constrained optimization problem defined by (5-6)\\
Initialize $\lambda=\lambda_{0}$, $itr=0$\\
\While{$itr < MaxItr$}{
  Rewrite $L(\theta, \lambda)$ to $\sum_{t=0}^{T-1} c(s_t, a_t)$\\
  Solve for optimal $\pi_\theta$ using LQR\\
  Evaluate constraint violation $\Delta = D_{KL}(p(\tau)\|p(\hat{\tau}))-\epsilon)$\\
  Update $\lambda \leftarrow \lambda+\alpha\Delta$
}
\caption{Dual Gradient Descent for Policy Optimization}
\end{algorithm}

\section{Experimental Setup and Baselines}
\subsection{The Carla Urban Driving Simulator and Scenarios}

We conducted our experimental validation using the Carla urban driving simulator, which is a high-definition open-source simulator by Intel. We conducted experiments under three different scenarios, straight lane, $90\degree$ turning, and circular roundabout. For each scenario, we investigated the performance of the proposed method and a series of baseline algorithms with and without obstacle vehicles. The Carla simulator and the simulated scenarios are shown in Fig. 2. In order to test the performance of the RL methods serving as motion planner, we directly extracted the vehicles and map states from the Carla simulator, and designed the state input for RL policy. Because the RL policy take in fixed dimensional states, we designed two kinds of states for the scenarios with and without obstacles. For the tasks without obstacles, the state input include the lateral deviation and yaw error of the autonomous vehicle with respect to the road block, and the velocity of the autonomous vehicle. For the cases with obstacles, in order to maintain fixed state dimension, we investigate only the influence of the front vehicle, and augment the state of the relative position and velocity between the autonomous vehicle and the front vehicle to the previously defined state, to obtain the state for these cases. We also designed concise and effective cost functions to model the objective of the driving tasks. For the road block tracking without considering the obstacle, we define the tracking cost to be

\begin{align}
c_t(s_t, a_t) = \alpha_l \Delta y^2 + \alpha_y \Delta \phi^2 + \alpha_v (v-v_{ref})^2 + \alpha_a a^2 + \alpha_\sigma \sigma^2
\end{align}

where $\Delta y$ is the lateral deviation, $\Delta \phi$ is the yaw angle error, $v$ is the velocity of the autonomous vehicle, $v_{ref}$ is the reference speed for the tracking, $a$ is the acceleration action, and $\sigma$ is the steering action. When considering the obstacle, we designed a nonlinear cost function that only takes effect when the autonomous vehicle is within the same lane of the front vehicle, and the distance is smaller than 20m, where we add the additional term of

\begin{align}
c_{aug}(s_t, a_t) = \beta_s (20-s) + \beta_v(v-v_{front})
\end{align}
where $v_f$ is the speed of the front vehicle.

For the GPS architecture, we used a GMM of 20 mixtures to serve as the model prior, and collect 4 trajectories every time for the update of the system dynamics and the linear Gaussian policy. The parameters for the cost function are: $\alpha$ We compare the proposed GPS based model based RL method with a series of popular model based and model free RL methods, in terms of their training and testing performance. We described the setting for the proposed method and briefly introduce the baseline algorithms and their settings in the following subsections.

\begin{figure}[t]
\centering
\includegraphics[scale=0.28]{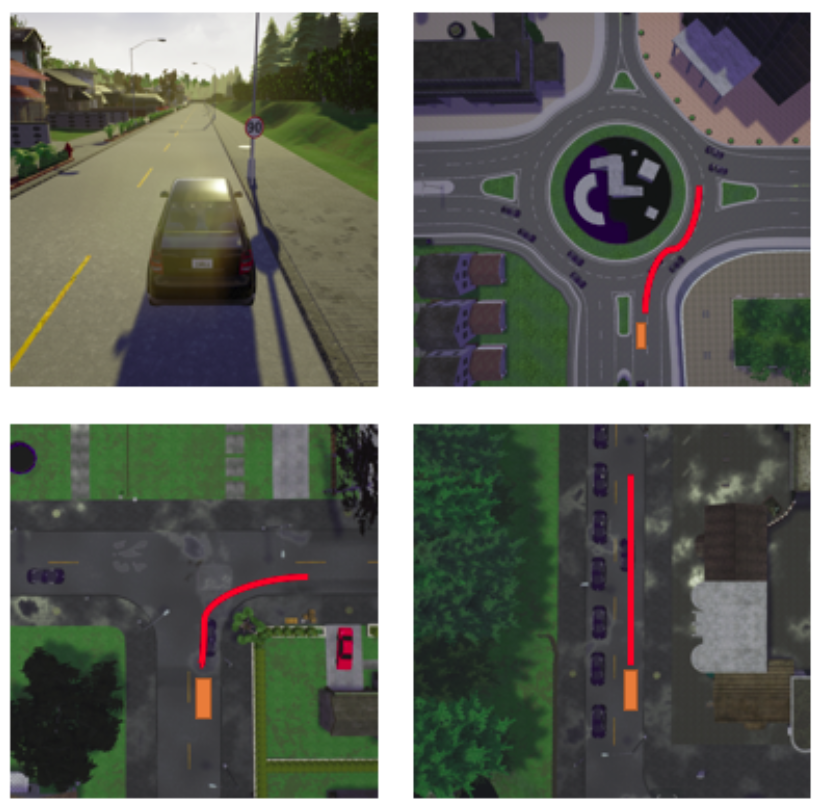}
\caption{The Carla urban driving simulator. Upper Left: the ego view of the autonomous vehicle; Upper Right: the roundabout experiment setting; Lower Left: the 90$\degree$ turning experiment setting; Lower Right: the straight driving experiment setting.}
\label{fig:structure}
\end{figure}

\subsection{Cross Entropy Policy Search RL Method}

Cross entropy method (CEM) \cite{de2005tutorial} has been one of the most simple and popular policy search RL method for policy optimization. In order to optimize the parameterized policy $\pi_\theta$, the CEM adopts the assumption of Gaussian distribution of $\theta = \mathcal{N} (\mu, \sigma^2)$. It iteratively samples $\theta$ from the distribution, using which to collect sample trajectories, and then updates $\mu$ and $\sigma$ using the $\theta$'s that produces the best trajectories. In our implementation, since the CEM is also a policy search method and can be sample efficient, we also collect 4 trajectories every time for policy update.

\subsection{Soft Actor Critic Model Free RL Method}

The soft actor critic (SAC) \cite{haarnoja2018soft} algorithm is currently the state of the art model free RL algorithm, enjoying the best sample complexity and the convergence property. The SAC maximizes both the expected reward (negative cost) and the entropy

\begin{align}
\max_\theta \sum_{t=0}^{T-1} E_{\pi_\theta} [-c(s_t, a_t) + \alpha H(\pi_\theta)]
\end{align}

The SAC adopts soft Bellman backup operator to solve for the optimal soft Q value function, resulting in a stochastic neural network policy. In our implementation, the policy network and the critic network are both fully connected neural networks with two hidden layers of 256 neurons. Each time step, we collect a batch size of 256 steps for the off policy model free RL training, and adopt a learning rate of 0.0003.

\begin{figure}[t]
\centering
\includegraphics[scale=0.5]{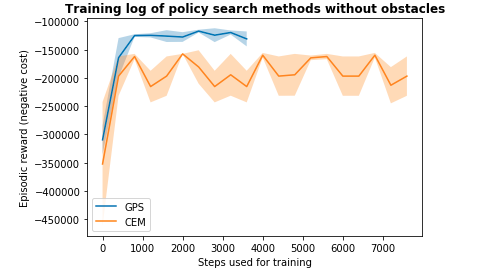}
\caption{The training log of the guided policy search (GPS) and cross entropy methods (CEM) on the driving task without obstacles. The GPS based method converges faster and to a better driving policy compared to the CEM.}
\label{fig:structure}
\end{figure}

\begin{figure}[t]
\centering
\includegraphics[scale=0.5]{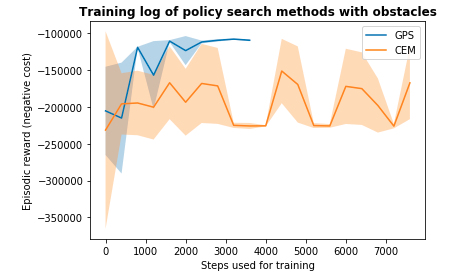}
\caption{The training log of the guided policy search (GPS) and cross entropy methods (CEM) on the driving task with obstacle. Similar to the task without obstacle, the GPS based method converges faster and to a better driving policy compared to the CEM.}
\label{fig:structure}
\end{figure}

\section{Experiment Results and Discussions}

The experiments are divided to two parts, the driving without obstacles and with obstacles, the two tasks have different dimension of input and different cost functions, as described in the previous section. For both cases, the autonomous vehicle is randomly initiated from one of the three settings, the straight driving, 90$\degree$ turning and roundabout entering. We run the proposed GPS based method and the baselines of CEM, SAC methods to learn the urban driving policies, and the figures below show the training log of the methods in comparison. For the initialization of the GPS and CEM, since the policies are linear Gaussian, we apply a PD controller as the initialization, with large variance, while for the model free RL, the policies are neural networks, so we follow the pure random initialization. Therefore, the initial performances of the GPS and CEM are slightly better compared to the model free RL methods.

Fig. 3 shows the training log of the policy search methods, the GPS and CEM methods on driving tasks without obstacles and Fig. 4 shows the training log of the two methods in tasks with obstacles. It is shown that in both cases, the GPS method outperforms the CEM in terms of both the speed and the optimum of convergence. The GPS algorithms converges with only about 1000 steps of data, corresponding to 100 seconds driving, while the CEM takes roughly twice as much time to converge. In terms of the qualitative performance, the GPS policy (roughly -100k reward) can track the road lane more stably and can surpass the front vehicle when it affects the speed profile of the autonomous vehicle, while the CEM methods (roughtly -200k reward) can make uncomfortable steering actions, and also sometimes collide with the obstacle or drive out of the road.

Fig. 5 shows the comparison of the model based and policy search methods with the state of the art model free RL method of SAC. Indeed the model free RL, as reported in the literature can achieve superior performance. In our experiment, we also report the model free method achieved a similar performance compared with our proposed GPS based method. However, the sample efficiency property of our proposed model based method is 100 times better, since the SAC takes more than 100k steps to converge.

\begin{figure}[!htb]
\minipage{0.122\textwidth}
  \includegraphics[width=\linewidth]{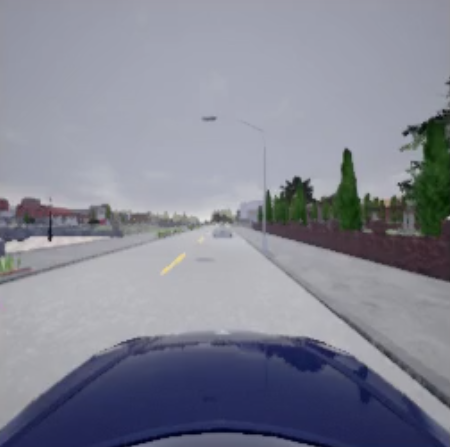}
\endminipage\hfill
\minipage{0.122\textwidth}
  \includegraphics[width=\linewidth]{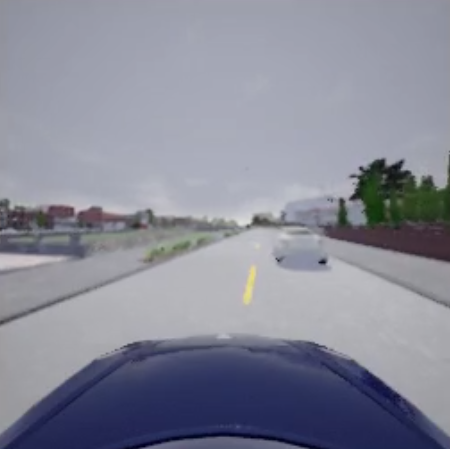}
\endminipage\hfill
\minipage{0.1215\textwidth}%
  \includegraphics[width=\linewidth]{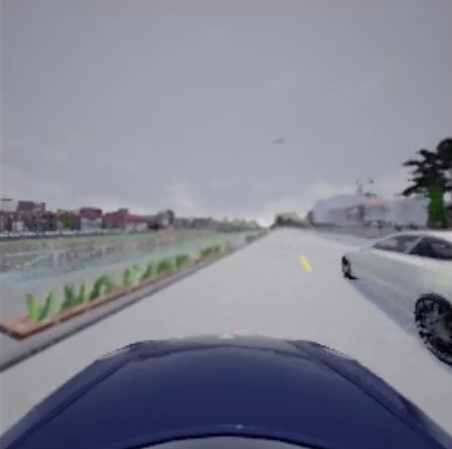}
\endminipage
\minipage{0.12\textwidth}%
  \includegraphics[width=\linewidth]{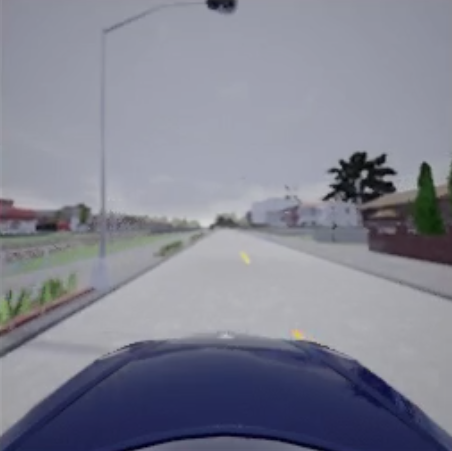}
\endminipage
\caption{Qualitative performance of the GPS policy of a case study of the task of actively changing lane to surpass the obstacle vehicle. This task involves high level decision and planning intelligence, which is hard to learn. In our study, only the proposed GPS method learned satisfactory policy for this case. Using CEM, IL and model free RL, this kind of task has not been solved in our previous works or in this work.}
\end{figure}

\begin{figure}[t]
\centering
\includegraphics[scale=0.5]{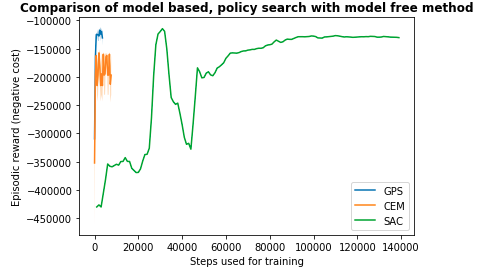}
\caption{The training log of the guided policy search (GPS) and cross entropy methods (CEM) in comparison with the soft actor critic (SAC) model free RL method on tasks without obstacles. The model free RL can learn satisfying driving policy, but it takes mode than 100k steps of data for training. That is, the model based method is 100x more sample efficient.}
\label{fig:structure}
\end{figure}

\section{Conclusion}

In this paper we proposed to use the model based RL method to learn urban autonomous driving policies. The effectiveness of the IL and model free RL in the real world application is polluted by their low sample efficiency, lack of interpretability, and difficulty for transfer. The model free RL normally requires millions of steps interaction with the environment. Therefore, we adopt the model based idea and uses a GMM to first approximate the system dynamics and then use a dual gradient descent to optimize the contrained policy optimization problem, subject to a trajectory change magnitude constraint. This variant of the so called guided policy search method is proved to be much more sample efficient compared to the model free RL methods, and also more effective compared to other popular policy search method of cross entropy method. Also, since the model learned using GMM has a clear interpretation in real world, it is beneficial when the policy is to be transferred to the real world, where the system dynamics needs to be adapted and can be adapted quickly. It is noted that currently the method has the weakness of it cannot dynamically manage different dimension of input, and future work include using a latent representation to directly uses raw sensor data to learn a universal policy based on model based RL.

\addtolength{\textheight}{-0cm}   
\bibliography{reference}
\bibliographystyle{ieeetr}

\addtolength{\textheight}{+0cm}
\newpage

\end{document}